\documentclass[runningheads]{llncs}
\usepackage[T1]{fontenc}
\usepackage{type1cm}
\usepackage{amsmath, amssymb}
\usepackage{mathrsfs}
\usepackage{multirow}
\usepackage{booktabs}
\usepackage{subfigure}
\usepackage{marvosym}

\usepackage{graphicx}

\begin{document}
\title{HIFI: Anomaly Detection for Multivariate Time Series with High-order Feature Interactions}
\titlerunning{HIFI}

\author{Liwei Deng\inst{1}\and
Xuanhao Chen\inst{1}\and
Yan Zhao\inst{2}\and
Kai Zheng\inst{1}\textsuperscript{(\Letter)}}
\authorrunning{L. Deng et al.}
\institute{School of Computer Science and Engineering, University of Electronic Science and Technology of China \\
\email{deng\_liwei@std.uestc.edu.cn, xhc@std.uestc.edu.cn, zhengkai@uestc.edu.cn} \and  Aalborg University, Danmark\\
\email{yanz@cs.aau.dk}}

\maketitle              
\begin{abstract}
Monitoring complex systems results in massive multivariate time series data, and anomaly detection of these data is very important to maintain the normal operation of the systems. Despite the recent emergence of a large number of anomaly detection algorithms for multivariate time series, most of them ignore the correlation modeling among multivariate, which can often lead to poor anomaly detection results. In this work, we propose a novel anomaly detection model for multivariate time series with \underline{HI}gh-order \underline{F}eature \underline{I}nteractions (HIFI). More specifically, HIFI builds multivariate feature interaction graph automatically and uses the graph convolutional neural network to achieve high-order feature interactions, in which the long-term temporal dependencies are modeled by attention mechanisms and a variational encoding technique is utilized to improve the model performance and robustness. Extensive experiments on three publicly available datasets demonstrate the superiority of our framework compared with state-of-the-art approaches.

\end{abstract}
\section{Introduction}
Complex systems such as servers \cite{OmniAnomaly} and aircrafts \cite{LSTM-NDT} are ubiquitous in the real world. Monitoring the behaviors of these systems generates huge amounts of multivariate time series data. A key task in managing complex systems is to detect system anomalies in a timely and accurate manner in order to reduce or avoid the losses caused by system anomalies. Due to the various of anomalies and lack enough labelled anomalies data, supervised anomaly detection methods are hard to adopt. In this work we will design an unspervised model to detect system anomalies.

Recently, many unsupervised anomaly detection models \cite{OED,OmniAnomaly,EncDec-AD} are proposed. EncDec-AD \cite{EncDec-AD} uses autoencoder architecture with LSTM and treats the reconstruction errors as anomaly scores to detect anomaly. OmniAnomaly \cite{OmniAnomaly} adopts advanced variational techniques to improve the ability of modeling complex time series. Although these methods can get better performance than their baselines for multivariate time series, they have some disadvantages.

Firstly, these studies ignore the correlation between multivariate, which is helpful for modeling complex temporal information \cite{MTGNN}. Secondly, they generally adopt RNN or its variants to model temporal information, which hardly capture the long-term temporal dependencies \cite{DA-RNN}. Thirdly, some of them use deterministic models \cite{LSTM-NDT} which are unrobustness and weak representation ability \cite{OmniAnomaly}.

To address the above-mentioned problems, we design a novel unsupervised model called HIFI (anomaly detection for multivariate time series with \underline{HI}gh-order \underline{F}eature \underline{I}nteractions). Specifically, a feature interaction graph is constructed automatically and then is delivered to GNN to model high-order feature interaction. To capture long-term dependencies and improve the robustness, attention-based time series modeling module and variational technique are used.

The main contributions of this paper are as follows:
\begin{itemize}
\item We design a multivariate feature interaction module, which uses the graph convolutional neural network to conduct high-order feature interactions on multi-dimensional temporal variables. 
\item We utilize the attention mechanism to model the long-term temporal dependence and variational encoding to improve the robustness of the model, which is critical for anomaly detection.
\item We conduct extensive experiments on three publicly available datasets, which empirically demonstrate the advantages of our proposed model compared with the representative models.
\end{itemize}

\section{Related Work}
In this section, we will introduce some work related to our model, such as unsupervised time series anomaly detection, graph convolutional neural network, and time series modeling method based on attention mechanism.

EncDec-AD is proposed by \cite{EncDec-AD}, which adopts the structure of sequence to sequence and uses two independent LSTM models as encoder and decoder respectively. \cite{LSTM-NDT} develops the LSTM-NDT model, which is used to model the temporal information through LSTM. OmniAnomaly \cite{OmniAnomaly} aims to capture the normal patterns of multivariate time series by stochastic variable connections and panar normalizing flow. OED \cite{OED} is developed to use recurrent autoencoder ensembles to detect anomaly. However, none of them explicitly model the relationship between features. Therefore, in this work, we use the graph neural network to model the high-order interaction features.

Graph Convolutional network (GCN) is firstly proposed in \cite{GCN}, which shows a strong representation ability of unstructured graph in node classification task and attracts the attention of a large number of researchers. Then a series of variants of GCN are proposed. GAT \cite{GAT} introduces the correlation between nodes through the attention mechanism. PPNP \cite{PPNP} uses personalized pagerank to extend the size of neighborhood and avoids oversmooth.

Transformer \cite{Transformer} is the first time series modeling approach that completely abandons both recurrent nerual network structures and convolutional nerual network structures. Its performance reflects the superiority of its model structure. SASRec \cite{SASRec} also adopts the attention mechanism to model the sequential relation of items, and achieves the good performance in the field of sequential recommendation. However, few works adopt the attention mechanism in the field of anomaly detection.

\begin{figure}[t]
\centering
\includegraphics[width=0.9\columnwidth]{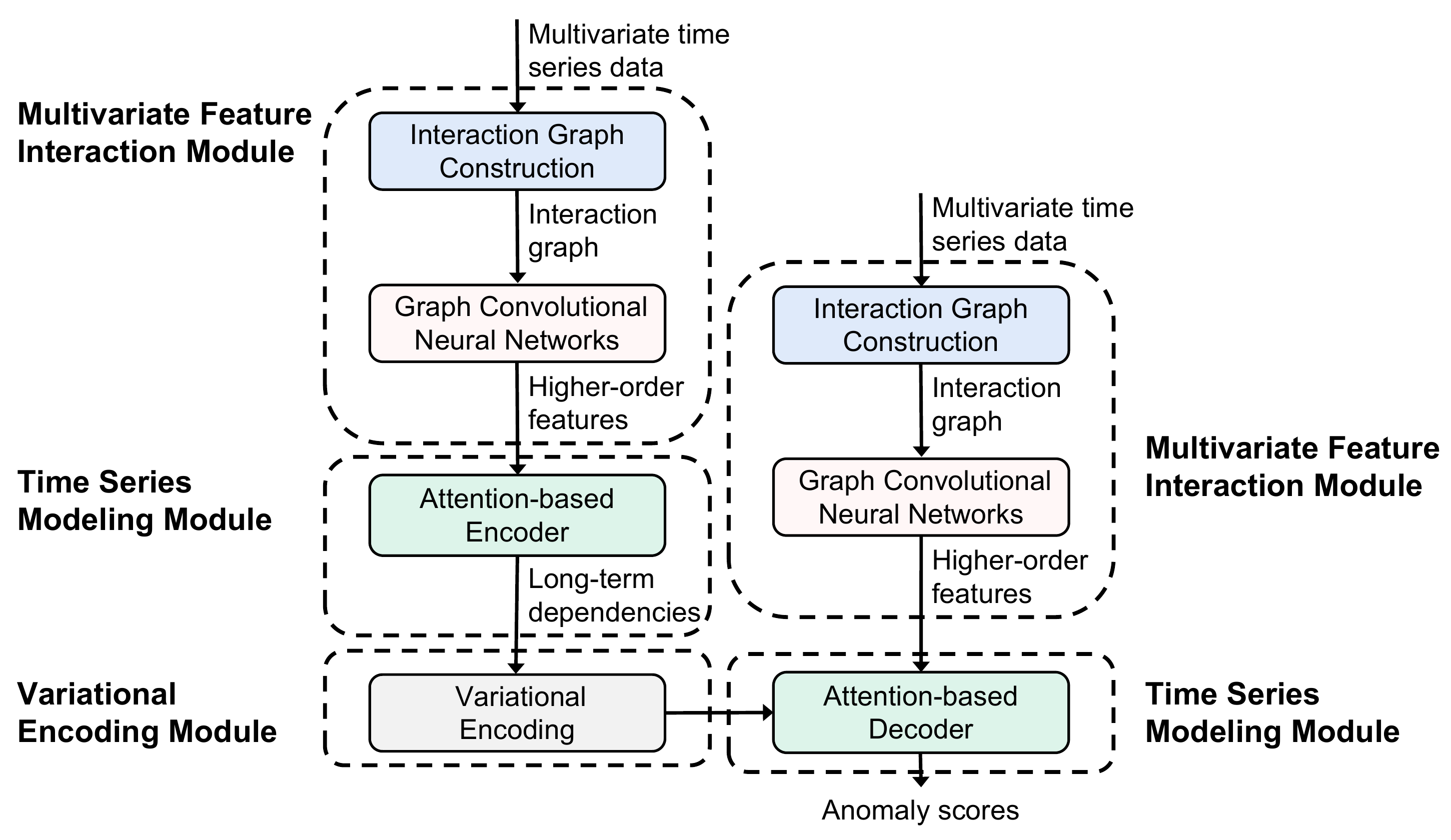} 
\caption{Framework overview. Our model adopts encoder-decoder architecture in which three modules are involved such as multivariate feature interaction module, attention-based time series modeling module and variational encoding module.}
\label{fig1}
\vspace{-0.6cm}
\end{figure}
\vspace{-0.3cm}
\section{Methodology}
As shown in Figure \ref{fig1}, our model consists of three type of parts, namely, the multivariate feature interaction module, the attention-based time series modeling module, and the variational encoding module. Each part will be elaborated in the following.
\vspace{-0.3cm}
\subsection{Multivariate Feature Interaction Module}
In order to model the relationship between variables, the multivariate feature interaction module firstly constructs the interaction graph through feature embedding, and then gets the high-order interaction features through the graph convolutional neural network. In practice, it is hard to use graph convolutional neural network to get high-oder features for multivariate time series because of the agnostic of relation graph among multivariate. So we construct the relation graph automatically in which each node represents a variable.

Inspired by MTGNN \cite{MTGNN}, we directly transform the original features $X=(x_{t-w+1}, $ $x_{t-w+2}, \cdots, x_{t})\in \mathbb{R}^{w \times d}$ to the hidden space $X^h=(x_{t-w+1}^h, x_{t-w+2}^h,$ $ \cdots, x_{t}^h) \in \mathbb{R}^{w \times d_1}$. In this way, we can control the size of the interaction graph. When the original features have a higher dimension, we can set a smaller $d_1$ to reduce the amount of calculation of the graph convolution.
\begin{equation}
\footnotesize
x_{i}^h = W_h x_i + b_h
\label{igc1}
\end{equation}
where $W_h \in \mathbb{R}^{d_1 \times d}$, $b_h \in \mathbb{R}^{d_1}$ are model parameters.

Then, in order to obtain the interaction graph with more expressive capability, we adopt an asymmetric construction method and define two independent feature embedding matrices $E_1$ and $E_2$, in which the $i$-th row of each matrix represents the embedding of $i$-th feature. Then we calculate the embedding similarity of each feature, and express the strength of feature correlation by the similarity as follow.
\begin{equation}
\footnotesize
\begin{aligned}
M_1= \ &tanh(E_1 \Theta_1) \\
M_2= \ &tanh(E_2 \Theta_2) \\
A=ReLU(tanh&(M_1 M_2^T - M_2^T M_1))
\end{aligned}
\end{equation}
where $E_1 \in \mathbb{R}^{d_1 \times d_1}$, $E_2 \in \mathbb{R}^{d_1 \times d_1}$, $\Theta_1 \in \mathbb{R}^{d_1 \times d_2}$, $\Theta_2 \in \mathbb{R}^{d_1 \times d_2}$ are model weights, which are learnable during training. $A \in \mathbb{R}^{d_1 \times d_1}$ is the feature interaction graph.

Obviously, the interaction graph constructed by the above methods is a complete graph. In order to reduce the computation of graph convolution module, $topk$ is used to return the maximum $k$ values in each row of the adjacency matrix, turning the complete graph into a sparse graph.
\begin{equation}
\footnotesize
\begin{aligned}
\mathbb{A}=\ &topk(A) \\
\end{aligned}
\vspace{-0.3cm}
\end{equation}

\subsubsection{Graph Convolution Module} We take the constructed feature interaction graph $\mathbb{A}$ and the hidden features $X^h$ into the graph convolutional network \cite{PPNP} to obtain the higher-order interaction features. We first calculate $\hat{A} = \tilde{D}^{-\frac{1}{2}} \tilde{A} \tilde{D}^{-\frac{1}{2}}$, where $\tilde{A} = \mathbb{A} + I_N$ and $I_N$ is identity matrix and $\tilde{D}_{ii} = \sum_j \tilde{A}_{ij}$.
\begin{equation}
\footnotesize
H_{k+1} = (1-\alpha)\hat{A}H_k + \alpha H_0
\end{equation}
where $H_k \in \mathbb{R}^{d_1 \times w}$ represents the high-order interaction features obtained by the $k$-th convolution. $H_0$ is the hidden features ${X^{h}}^T$. $\alpha$ is a hyperparameter defining the amount of information that retains the original feature at every convolution. It is important to note that the feature interaction does not interact across time steps. The higher-order intereaction features of the current time step are completely calculated by its own hidden features.

Moreover, since the convolution depth required by each high-order feature may be different, we concatenate the output of the convolution of each step to obtain the final higher-order feature representation through linear transformation. 
\begin{equation}
\footnotesize
X^{ho} = Concat(H_0^T, H_1^T, ..., H_K^T)W^{ho}
\end{equation}
where $X^{ho}=(x_{t-w+1}^{ho}, x_{t-w+2}^{ho}, \cdots, x_{t}^{ho}) \in \mathbb{R}^{w \times d_1}$ is the high-order features corresponding to $X^h$. $Concat$ denote the concatenation operation. $W^{ho}\in \mathbb{R}^{Kd_1 \times d_1}$ is the model weights. $K$ is the maximal graph convolutional depth.
\vspace{-0.3cm}
\subsection{Attention-based Time Series Modeling Module}
We use the attention mechanism to model temporal information and capture long-term temporal dependencies, which are hardly captured by RNN model.
\vspace{-0.5cm}
\subsubsection{Attention Layer} We use one of the most common attention mechanisms, scaled-dot attention \cite{Transformer}, which can be described as follows.
\begin{equation}
\footnotesize
Attention(Q, K, V) = softmax(\frac{QK^T}{\sqrt{d_k}})V
\end{equation}
where $Q \in \mathbb{R}^{w \times d_k}$, $K \in \mathbb{R}^{w \times d_k}$, $V \in \mathbb{R}^{w \times d_v}$ are queries, keys and values respectively. $\sqrt{d_k}$ is the scaling factor. Intuitively, the nature of the attention mechanism is to take the weighted sum of all time steps according to the weights calculated. It allows information from any distance of time steps to flow directly to the current step, giving the attention mechanism the ability to capture long-term temporal dependencies. Multi-head attention \cite{Transformer}, which can allow the model to jointly attend to information from different representation subspaces at different positions, is used in our work.
\vspace{-0.3cm}
\subsubsection{Nonlinear Layer} Although multi-head attention can aggregate the information of each time step through adaptive weights, it is still a linear model. This greatly limits the capability of the model, so we use two linear network with $ReLU$ activation to introduce nonlinear information to the model.
\begin{equation}
\footnotesize
NonLinear(X) = W_2^{n}ReLU(W_1^{n}X+ b_1^{n}) + b_2^{n}
\end{equation}
where $W_1^{n} \in \mathbb{R}^{d_3 \times d_1}$, $W_2^{n} \in \mathbb{R}^{d_1 \times d_3}$, $b_1^{n} \in \mathbb{R}^{d_3}$, $b_2^{n} \in \mathbb{R}^{d_1}$ are model parameters. Due to the nature of position invariance of attention layer, we follow \cite{Transformer} to introduce sinusoidal positional encoding $P\in \mathbb{R}^{w*d_1}$ to add the sequential information after multivariate feature interaction module. Specifically, $X^{in}=P+X^{ho}+X$ where $X^{in}$ is the input of attention-based time series modeling module. Our encoder and decoder can be obtained by alternately stacking multi-head attention layer and nonlinear layer, in which layer normalization and residual connection are used to prevent overfitting and gradient disappearance respectively. 

\vspace{-0.3cm}
\subsection{Variational Encoding Module}
To improve the robustness and performance of the model, we model the deterministic encoding of the encoder $X^{eo}=(x_{t-w+1}^{eo}, x_{t-w+2}^{eo}, \cdots, x_{t}^{eo}) \in \mathbb{R}^{w \times d_1}$ as a normal distribution. We obtain the mean and the logarithm of variance of the normal distribution by two independent linear layers.

\begin{equation}
\footnotesize
\begin{split}
\mu_i = W_{\mu}x_{i}^{eo} + b_{\mu} \\
log \ \sigma^2_i = W_{\sigma}x_{i}^{eo} + b_{\sigma}
\end{split}
\end{equation}
where $W_{\mu} \in \mathbb{R}^{d_1 \times d_1}$, $W_{\sigma} \in \mathbb{R}^{d_1 \times d_1}$, $b_{\mu} \in \mathbb{R}^{d_1}$, $b_{\sigma} \in \mathbb{R}^{d_1}$ are model weights. 
We then adopt the reparameterization trick to sample from the normal distribution and input the samples into the decoder. We fix the number of samples at 1.
\begin{equation}
\footnotesize
z_i = \mu_i + \epsilon * \sigma_i
\end{equation}
where $\epsilon$ is a sample from $N(0,1)$. 
$\mu_i$ and $\sigma_i$ are the mean and the variance of normal distribution in $i$-th time step. The resampled $Z$ is inputted to attention-based time series modeling module of decoder and treated as $K$ and $V$.

\vspace{-0.5cm}
\subsection{Model Training}
\vspace{-0.3cm}
We use the reconstruction error of the entire current window and the Kullback-Leibler divergence between variational encoding of sample and the standard normal distribution to train the model.
\begin{equation}
\footnotesize
\label{eq2}
loss = \sum_{i=1}^w||x_i - x_{i}^{rec}||_2 + \beta \sum_{i=1}^w KL(N(\mu_i, \sigma_i^2)||N(0,1)))
\end{equation}
where $\beta$ is a hyperparameter, which is used to balance the loss of the two parts. 

In current window, we regard $||x_t-x_{t}^{rec}||_2$ as the anomaly score, based on which we can detect anomalies.
\vspace{-0.5cm}
\section{Experiments}
\vspace{-0.3cm}
In this section, we evaluate our model by comparing it with some state-of-the-art models. We begin by introducing the setup of the experiment, and then report and analyze the results of the experiment. 
\vspace{-0.4cm}
\subsection{Experimental Setup}
\subsubsection{Datasets and Metrics} 
We conduct experiments on three publicly available datasets, i.e. SMD (Server Machine Dataset) \cite{OmniAnomaly}, SMAP (Soil Moisture Active Passive satellite) and MSL (Mars Science Laboratory rover) \cite{LSTM-NDT}, in two different domains. We use Precision($Pre$), Recall($Rec$) and F1-score($F1$) to evaluate the performance of HIFI and baselines. We enumerate all possible anomaly thresholds to search for the best F1, denoted as $F1_{best}$. And a point-adjust \cite{KPI} is adopted to get the final prediction which is same as \cite{OmniAnomaly}.
\vspace{-0.5cm}
\subsubsection{Baselines} 
In our experiments, we select four representative models. LSTM-NDT \cite{LSTM-NDT} predicts the values of time step $t$ and uses predictive error as anomaly scores of step $t$. EncDec-AD \cite{EncDec-AD} adopts autoencoder architecture with LSTM and treats reconstruction error as anomaly score. OED-IF \cite{OED} employs multiple autoencoder with different connection structures to improve the performance of anomaly detection. OmniAnomaly \cite{OmniAnomaly} adopts advanced variational encoding techniques with autoencoder architecture to detect anomaly.
\vspace{-0.5cm}
\subsubsection{Implementation details} 
All models take the sliding window data of the original data as input and we set the window size $w$ to 100. We randomly select 30\% from the training data as validation sets, set the batch size as 64 for training and run for 100 epochs. We save the model with the least loss of the validation set as the final test model. We use Adam optimizer for stochastic gradient desent with an initial learning rate of 0.005. For our model, we turn hyperparameters in validation set. Specifically, we set $d_1=64$, $d_2=64$, $d_k=16$, $l=2$, $\alpha=0.2$, $\beta=1$, $K=3$ for all datasets. In MSL dataset, we set $d_3=256$ and in other datasets, we set $d_3=128$. For all of the baselines, if they are tested on the same dataset, we follow the settings of the original paper, or we tune the model to be optimal.

\subsection{Overall Performance Comparison}
\begin{table}[t]
\centering
\caption{The performance of HIFI with other baseline methods over three datasets}\smallskip\smallskip
\label{overall_performance}
\renewcommand\tabcolsep{3.0pt}
\resizebox{1.\columnwidth}{!}{
\begin{tabular}{c c c c c c c c c c}
\toprule
\multirow{2}{*}{Methods} & \multicolumn{3}{c}{MSL} & \multicolumn{3}{c}{SMAP} & \multicolumn{3}{c}{SMD} \\
& $F1_{best}$ & $Pre$ & $Rec$ & $F1_{best}$ & $Pre$ & $Rec$ & $F1_{best}$ & $Pre$ & $Rec$ \\
\midrule
LSTM-NDT & 0.8623 & 0.7830 & 0.9594 & 0.7852 & 0.6756 & 0.9373 & 0.7942 & 0.6865 & 0.9481 \\
EncDec-AD & 0.9039 & 0.8606 & 0.9520 & 0.8707 & 0.7737 & 0.9956 & 0.9491 & 0.9317 & 0.9673  \\
OED-IF & 0.9185 & 0.8754 & 0.9661 & 0.8458 & 0.7351 & \textbf{0.9959} & 0.9730 & 0.9685 & \textbf{0.9777} \\
OmniAnomaly & 0.9257 & 0.8802 & 0.9762 & 0.8966 & 0.8198 & 0.9893 & 0.9503 & 0.9337 & 0.9675 \\
HIFI & \textbf{0.9546} & \textbf{0.9133} & \textbf{0.9998} & \textbf{0.9708} & \textbf{0.9475} & 0.9952 & \textbf{0.9773} & \textbf{0.9811} & 0.9737 \\
\bottomrule
\end{tabular}
}
\vspace{-0.3cm}
\end{table}

Table \ref{overall_performance} shows the performance of our model compared with the baseline models in three datasets. LSTM-NDT performs worst in all the baselines in terms of $F1_{best}$, which shows the disadvantage of the predictive model. In most cases, OED-IF can get better performance than EncDec-AD, which shows that model ensemble is a good strategy for anomaly detection. OmniAnomaly generally performs better than other baselines, which illustrates the variational encoding can improve the performance of anomaly detection. Last but not least, HIFI outperforms all baselines on three datasets in terms of $F1_{best}$. In particular, HIFI outperforms the best-performing state-of-the-art method by 2.89\%, 7.42\% and 0.43\% on MSL, SMAP and SMD dataset respectively, which shows the effectiveness of the proposed model.
\vspace{-0.3cm}
\subsection{Ablation Study}
We perform an ablation study at $F1_{best}$ to verify the effectiveness of each component of our model. We name different variants of HIFI as follows:

\textbf{w/o FI: } HIFI without multivariate feature interaction module.

\textbf{w/o VE: } HIFI without variational encoding module.

\textbf{w/o FI+VE: } HIFI without multivariate feature interaction module and stochastic variable embedding module.

\textbf{w/o FI+VE+EN: } HIFI only with encoder part. We stack 4 encoder layers to fair comparison.

The results of our model and its variants are shown in the Table \ref{tab3}.
In most cases, HIFI can achieve the best results. In the SMD dataset, the performance of each model is similar. This is because the time series information is sufficient to detect anomalies, and introducing complex features does not improve the performance of the model. In MSL and SMAP datasets, we can clearly see that w/o FI and w/o VE outperform w/o FI+VE, which proves the effectiveness of the multivariate feature interaction module and variational encoding module. Comparing performance between w/o FI+VE and w/o FI+VE+EN, we can find that w/o FI+VE can get better performance, which shows the superiority of Encoder-Decoder model in anomaly detection. Finally, compared with the baselines, w/o FI+VE+EN can get better performance generally, which shows the importance of long-term temporal dependency modeling in anomaly detection.

\begin{table}[t]
\vspace{-0.3cm}
\centering
\caption{The $F1_{best}$ of HIFI and its variants}\smallskip
\label{tab3}
\resizebox{0.5\columnwidth}{!}{
\begin{tabular}{c c c c}
\toprule
Methods & MSL & SMAP & SMD \\
\midrule
HIFI & \textbf{0.9546} & \textbf{0.9708} & 0.9773 \\
w/o FI & 0.9455 & 0.9674 & 0.9781 \\
w/o VE & 0.9423 & 0.8821 & 0.9777 \\
w/o FI+VE & 0.9415 & 0.8703 & \textbf{0.9787} \\
w/o FI+VE+EN & 0.9358 & 0.8295 & 0.9771 \\
\bottomrule
\end{tabular}
}
\vspace{-0.4cm}
\end{table}

\section{Conclusion}
\vspace{-0.3cm}
In this paper, we propose a model, namely HIFI, an unsupervised anomaly detection model for multivariate time series with high-order feature interaction.  Extensive empirical study based on real datasets confirms the advantage of our proposed model compared with the state-of-the-art methods as well as the importance of different components of our model for multivariate anomaly detection. \\

\noindent
\textbf{Acknowledgements.} This work is supported by NSFC (No. 61972069, 61836007, 61832017) and Sichuan Science and Technology Program under Grant 2020JDTD0007.

\bibliographystyle{splncs04}
\bibliography{refs}

\end{document}